\documentclass{article} 
\usepackage{iclr2027_conference,times}

\usepackage{amsmath,amsfonts,bm}

\def\eqref#1{equation~\ref{#1}}

\def\1{\bm{1}}

\DeclareMathAlphabet{\mathsfit}{\encodingdefault}{\sfdefault}{m}{sl}
\SetMathAlphabet{\mathsfit}{bold}{\encodingdefault}{\sfdefault}{bx}{n}

\usepackage{hyperref}
\usepackage{url}

\usepackage{booktabs}
\usepackage{graphicx}
\usepackage{xcolor}
\usepackage{pifont}
\usepackage{amssymb}
\usepackage{multirow}
\usepackage{wrapfig}

\definecolor{CustomGreen}{RGB}{100,178,70}
\definecolor{CustomRed}{RGB}{235,35,35}
\definecolor{CustomYellow}{RGB}{230,170,0}

\newcommand{\cmark}{\textcolor{CustomGreen}{\ding{51}}}
\newcommand{\xmark}{\textcolor{CustomRed}{\ding{55}}}

\title{PReCache: Efficient KV Cache Sharing \\ for Multi-LoRA Agents via Low-Rank \\ Precomputation and Neutral Reconstruction}

\author{Hyesung Jeon, Hyeongju Ha, Jae-Joon Kim\\
Department of Electrical and Computer Engineering\\
Seoul National University\\
\texttt{\{hjeon2k,mnv1009,kimjaejoon\}@snu.ac.kr} \\
}

\iclrfinalcopy 
\begin{document}

\maketitle

\begin{abstract}
Multi-LoRA agent systems enable efficient role specialization by sharing a common backbone model.
However, each agent repeatedly processes the growing shared trajectory and constructs its own KV cache, introducing substantial memory and computation redundancy in long-horizon tasks.
Existing KV cache sharing methods reduce this repeated prefill, but they either require additional training or architectural constraints or retain substantial model computation.
Moreover, direct cache reuse causes the current agent to rely on cache states generated by the previous agent's adapter, weakening the role-specific behavior encoded by its own LoRA.
We present \textbf{PReCache}, a training-free KV cache sharing framework with two designs, namely \textbf{PreLRShared} and \textbf{ReBaseShared}, that share the base cache computed using the pretrained weights and precompute a compact agent-specific low-rank (LR) cache.
To remove repeated prefill, PreLRShared precomputes each agent's LR cache when the shared context is first processed, allowing the current agent to use its own LR cache without reprocessing context processed by previous agents.
To improve sharing accuracy, ReBaseShared reconstructs the shared base cache from adapter-free hidden states, reducing the remaining error caused by the previous agent's adapted representation.
To minimize its reconstruction cost, we propose two inference schemes tailored to single-stream inference and concurrent serving, performing the same reconstruction after each agent's turn or alongside its execution, respectively.
Across multiple models and agent benchmarks, PreLRShared achieves up to a $3.1\times$ TTFT speedup and a $2.3\times$ improvement in per-request throughput over inference without KV cache sharing. ReBaseShared best preserves accuracy overall among the evaluated cache-sharing methods, with an average drop of only $1.1\%$ points relative to inference without cache sharing.
\end{abstract}

\section{Introduction}
\label{sec:intro}
Large language models (LLMs) are widely deployed as agents that decompose complex tasks into multiple subtasks~\citep{xu2023rewoo,liu2023bolaa,li2023camel,hong2023metagpt,shen2023hugginggpt,wu2024autogen,qiao2024autoact}, invoke external tools to obtain observations~\citep{schick2023toolformer,qin2023toolllm,zhou2024webarena,yang2024sweagent,drouin2024workarena}, reflect on and revise their decisions~\citep{shinn2023reflexion,madaan2023selfrefine,gou2024critic,zhang2024selfcontrast}, and operate through iterative loops of model calls~\citep{yao2023react,zhou2023lats,liu2024agentbench,wang2025awm,chen2025longhorizon}.
Across these settings, agents are often assigned specialized roles~\citep{li2023camel,hong2023metagpt,wu2024autogen,qiao2024autoact,chung2026agentx}.
LoRA provides an efficient way to implement this specialization by sharing a pretrained backbone and using a small additive adapter fine-tuned for each role~\citep{hu2022lora,liu2024dora,sheng2024slora,shen2025edgelora,lee2026mapcoderlite,zeng2026marlworkflow}.
Recent methods further replace role-specific fine-tuning by directly converting role descriptions and agent prefixes into LoRA adapters, broadening the applicability of multi-LoRA agent systems~\citep{phang2023hypertuning,charakorn2025texttolora,liu2026shine,charakorn2026doctolora}.
By sharing the backbone weights across roles, multi-LoRA agent systems reduce model memory usage, which is particularly beneficial in resource-constrained settings~\citep{qiao2024autoact,li2025mobilora,fu2026minddrive,belcak2025slmagents,wang2025gamedialogue,shekar2025adaptiveminds}.

Despite sharing a common backbone, each multi-LoRA agent still processes the shared context independently, introducing substantial memory and computational redundancy.
Multi-agent systems accumulate context consisting of the user request, retrieved information, tool observations, and previous agents' outputs, collectively forming a shared trajectory~\citep{yao2023react,qiao2024autoact,zhuge2024gptswarm,zhang2025cut}.
As the number of agents and turns increases, this trajectory becomes increasingly long and prefill-heavy~\citep{kim2026characterization,huang2026pact,zhang2025cut}.
Each agent therefore repeats prefill over context already processed by previous agents and constructs a separate KV cache for the same context~\citep{bian2026tokendance}.
KV cache sharing removes this redundancy, but naively reusing KV caches generated under different adapter weights causes substantial accuracy degradation.
Existing methods mitigate this mismatch either by selectively recomputing critical layers or tokens~\citep{yao2025cacheblend,liu2026droidspeak,geng2026relaycaching} or through deviation correction~\citep{ye2025kvcomm,li2026graphflow,ma2026kamera}.
However, deviation correction methods target prefix-induced differences and provide no adapter-specific correction when prefixes are matched across agents.
Selective recomputation leaves accuracy degradation under heterogeneous LoRA adapters, while larger recomputation ratios incur substantial hidden state and KV cache computation~\citep{li2025mobilora,jeon2026lragent}.
LRAgent~\citep{jeon2026lragent} directly addresses KV cache sharing for multi-LoRA agents by decomposing the cache into a base cache and a compact agent-specific low-rank cache (LR cache).
Its BaseShared method shares the base cache while retaining an agent-specific LR cache, mitigating KV cache sharing error while substantially reducing KV cache memory usage.
However, each agent still needs to reprocess the shared context to construct its own LR cache, leaving most of the repeated prefill computation unreduced.
As a result, BaseShared incurs computational costs comparable to or even higher than those of token-wise recomputation methods, despite better preserving accuracy.
Approaches that eliminate this repeated processing instead require specific architectures and adapters trained accordingly, limiting their direct application to existing multi-LoRA agents~\citep{woo2026icarus,woo2026prefillshare,jeon2026lragent}.
Thus, reducing both KV cache memory usage and repeated prefill computation for existing multi-LoRA agents without substantial accuracy degradation remains a critical challenge.

In this paper, we present \textbf{PReCache}, a training-free KV cache sharing framework comprising two designs, \textbf{PreLRShared} and \textbf{ReBaseShared}, which address this challenge by precomputing compact agent-specific LR caches when each segment of the shared context is first processed.
Unlike BaseShared, PreLRShared constructs the compact LR caches of all agents when newly added context is first processed, allowing subsequent agents to use their own LR caches without reprocessing the accumulated trajectory.
We further find that the shared base cache constructed from adapter-free hidden states is closer to the current agent's base cache than that constructed from the previous agent's adapter-conditioned hidden states.
Based on this observation, ReBaseShared reconstructs the shared base cache from adapter-free hidden states, reducing its dependency on the previous agent while retaining PreLRShared's LR cache precomputation.
To optimize neutral reconstruction for different inference environments, we develop lazy prefill (LP) for single-stream inference and double batching (DB) for concurrent serving.
We theoretically analyze the reduction in base cache error and experimentally validate the resulting improvements in accuracy and serving efficiency.

\section{Background}
\label{sec:background}
\subsection{Multi-LoRA based agent systems}
\label{sec:background_multilora}

Multi-LoRA agent systems share a pretrained backbone across agents and use a lightweight LoRA adapter fine-tuned for each specialized role~\citep{qiao2024autoact,li2025mobilora,fu2026minddrive,wang2025gamedialogue,shekar2025adaptiveminds}.
We denote a frozen projection in the shared backbone by $W_0 \in \mathbb{R}^{d_{\mathrm{in}}\times d_{\mathrm{out}}}$ and the LoRA weights of agent $i$ by $A_i \in \mathbb{R}^{d_{\mathrm{in}}\times r}$ and $B_i \in \mathbb{R}^{r\times d_{\mathrm{out}}}$, where $r \ll d_{\mathrm{in}},d_{\mathrm{out}}$.
Given the adapter-conditioned hidden states $X_{i,\mathrm{adpt}} \in \mathbb{R}^{L\times d_{\mathrm{in}}}$ for a sequence of length $L$, the projection output is
\begin{equation}
    Y_i
    = X_{i,\mathrm{adpt}}(W_0 + A_iB_i)
    = \underbrace{X_{i,\mathrm{adpt}}W_0}_{\text{base contribution}}
    + \underbrace{(X_{i,\mathrm{adpt}}A_i)B_i}_{\text{adapter contribution}}.
\end{equation}
The adapter contribution $(X_{i,\mathrm{adpt}}A_i)B_i$ differs across agents, and these differences propagate through subsequent layers.
Consequently, the hidden states $X_{i,\mathrm{adpt}}$ become agent-dependent, causing even the base contribution $X_{i,\mathrm{adpt}}W_0$ to differ across agents.
Thus, although the agents receive the same shared context, each agent must process it with its own adapter to construct the corresponding KV cache.
This repeated processing increases computation, while maintaining a separate KV cache for each agent increases memory usage as the shared trajectory grows.

\subsection{Multi-Agent KV Cache Sharing}
\label{sec:background_sharing}

\begin{table}[t]
\centering
\vspace{-10pt}
\caption{Conceptual comparison of KV cache sharing strategies for multi-LoRA agents across three criteria. \cmark{} and \xmark{} denote support and no support, respectively.}
\label{tab:method_comparison}
\resizebox{1.0\linewidth}{!}{
\begin{tabular}{lccc}
\toprule
\textbf{Method}
& \textbf{Accuracy Preservation}
& \textbf{Repeated Prefill Reduction}
& \textbf{Existing Adapter Support} \\
\midrule
Selective Recomputation & \xmark & \cmark & \cmark \\
BaseShared               & \cmark & \xmark & \cmark \\
BaseLRShared             & \cmark & \cmark & \xmark \\
PReCache (Ours)          & \cmark & \cmark & \cmark \\
\bottomrule
\end{tabular}
}
\vspace{-10pt}
\end{table}

KV cache sharing across agents reduces the memory used for shared context and avoids repeated prefill for KV cache construction.
However, different adapter weights produce different KV caches even when agents process the same shared context.
Directly reusing the entire KV cache from the previous agent (\textbf{FullShared}) therefore causes accuracy degradation for the current agent.
Prior work on KV cache sharing across models or agents uses learned mappings, calibrated transformations, or specialized model structures to account for cache differences~\citep{fu2026c2c,dery2026latentalign,woo2026prefillshare,woo2026icarus,heo2026crossmodel}.
However, these methods require additional training, calibrated model pairs, or specific architectures, limiting their direct application to existing multi-LoRA agents.
Other training-free methods correct KV cache deviations caused by differences in prefixes, context relationships, or token positions~\citep{ye2025kvcomm,li2026graphflow,ma2026kamera}.
However, when prefixes and token positions are matched across agents, their correction variables remain unchanged and provide no adapter-specific correction.
Their unmodified application therefore reduces to FullShared for differences caused by LoRA adapters.

Selective recomputation instead recovers agent-specific KV caches by recomputing selected layers or tokens with the current agent and reusing the remaining cache.
DroidSpeak~\citep{liu2026droidspeak} identifies critical layer groups through offline profiling and processes all shared tokens through the selected layers, beginning from a stored hidden state before the first recomputed layer.
CacheBlend~\citep{yao2025cacheblend} selects tokens with high KV cache deviation and recomputes them through subsequent layers.
RelayCaching~\citep{geng2026relaycaching} further combines KV cache deviation with attention scores to select influential tokens within critical middle layers.
Overall, these methods prioritize the layers or tokens most sensitive to KV cache differences to recover accuracy within a limited recomputation budget.
However, heterogeneous LoRA adapters produce KV cache differences across multiple layers and tokens~\citep{li2025mobilora}.
A limited recomputation ratio therefore leaves accuracy degradation, whereas increasing the ratio to preserve accuracy recomputes and stores a larger portion of the current agent's KV cache, reducing both computation and memory savings.

LRAgent~\citep{jeon2026lragent} directly targets KV cache sharing for multi-LoRA agents by decomposing an adapted projection $Y_i$ into a base cache $X_{i,\mathrm{adpt}}W_0$ and an agent-specific LR cache $X_{i,\mathrm{adpt}}A_i$, following the notation in Section~\ref{sec:background_multilora}.
Its BaseShared method shares the base cache while maintaining a separate LR cache for each agent.
This decomposition preserves the agent-specific adapter contribution while reducing KV cache memory usage to a level close to that of FullShared.
However, to construct its LR cache, the current agent still performs full-length backbone processing over the accumulated shared context that it has not previously processed.
Consequently, BaseShared retains most of the repeated prefill computation and often incurs a computational cost comparable to or greater than that of token-wise recomputation, despite preserving accuracy more effectively.
BaseLRShared removes this repeated processing by using the same LoRA down-projection across agents, allowing a single LR cache to be shared.
However, all adapters must use the shared down-projection during training, so BaseLRShared does not directly support existing LoRA adapters with different down-projection weights.
Thus, preserving accuracy while reducing KV cache memory and eliminating most repeated prefill for existing multi-LoRA agents remains an open challenge.
Table~\ref{tab:method_comparison} summarizes these trade-offs.

\section{Methodology}
\label{sec:method}

\begin{figure*}[t]
    \centering
    \vspace{-15pt}
    \includegraphics[width=1.0\linewidth]{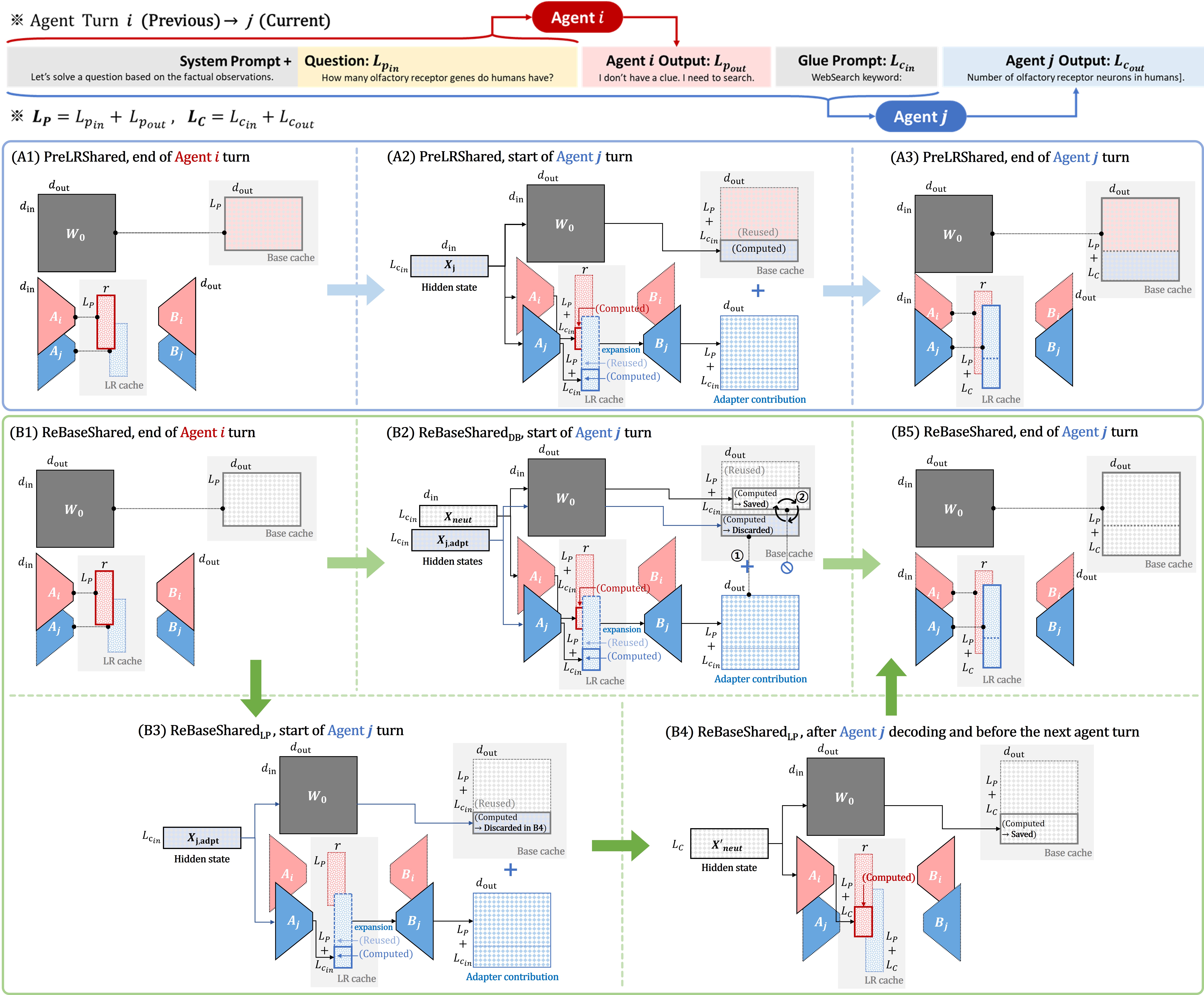}
    \vspace{-15pt}
    \caption{Overview of KV cache construction in PReCache across consecutive agent turns.
    (A1--A3) PreLRShared uses the hidden states generated when each context segment is first processed to construct the LR caches for all agents, eliminating the separate full-length backbone processing over the accumulated context $L_P$ required by BaseShared.
    (B1--B5) ReBaseShared reconstructs the shared base cache from adapter-free hidden states to reduce dependency on the previous agent's adapted representation.
    ReBaseShared$_{\mathrm{DB}}$ (B2) schedules the adapted and adapter-free paths together during both prefill and decoding, whereas ReBaseShared$_{\mathrm{LP}}$ (B3--B4) performs adapter-free reconstruction as a contiguous prefill after the current agent's turn.}
    \label{fig:method_overview}
\end{figure*}

We present the two designs of PReCache, PreLRShared and ReBaseShared, for efficient KV cache sharing across existing multi-LoRA agents.
Using the notation from Section~\ref{sec:background_multilora}, Figure~\ref{fig:method_overview} illustrates their KV cache construction across consecutive turns of \textbf{previous} agent $i$ and \textbf{current} agent $j$.
Agent $i$ processes the input context $L_{P_{\mathrm{in}}}$ through prefill and generates $L_{P_{\mathrm{out}}}$ through decoding.
We denote their concatenation by $L_P$, which has been processed by agent $i$ but not by agent $j$.
Agent $j$ then receives $L_P+L_{C_{\mathrm{in}}}$ as its input context and generates $L_{C_{\mathrm{out}}}$ through decoding.
We denote the context newly added during agent $j$'s turn, consisting of $L_{C_{\mathrm{in}}}$ and $L_{C_{\mathrm{out}}}$, by $L_C$, such that the shared trajectory grows from $L_P$ to $L_P+L_C$.
Parts (A1--A3) illustrate PreLRShared, which precomputes the LR caches for all agents when each context segment is first processed, eliminating the need for agent $j$ to reprocess $L_P$ to construct its LR cache.
Parts (B1--B5) illustrate ReBaseShared, which reconstructs the shared base cache from adapter-free hidden states to reduce dependency on the previous agent's adapted representation while retaining this LR cache precomputation.
The following subsections detail the cache construction and reuse procedures of both designs.

\subsection{PreLRShared: LR Cache Precomputation}
\label{sec:method_prelr}

In conventional multi-LoRA agent execution, each agent processes its entire input context and constructs a separate KV cache, which we refer to as \textbf{NonShared}.
For agent $j$, this requires processing $L_P + L_{C_{\mathrm{in}}}$ with its adapter even though agent $i$ has already processed $L_P$.
BaseShared reduces the resulting KV cache memory usage by sharing the base cache over $L_P$, but agent $j$ still processes $L_P$ to construct its own LR cache.
Specifically, constructing
$X_{j,\mathrm{adpt}}A_j \in \mathbb{R}^{L_P \times r}$
requires computing
$X_{j,\mathrm{adpt}} \in \mathbb{R}^{L_P \times d_{\mathrm{in}}}$
through the attention and MLP blocks of the backbone.
Thus, although the resulting LR cache has width $r$, its construction retains full-length backbone processing over $L_P$.

PreLRShared removes this repeated processing by constructing the LR caches for all agents when each context segment is first processed.
When agent $i$ processes $L_P$ in a system with $N$ agents, PreLRShared applies every agent's down-projection $A_k$ to the hidden states
$X_{i,\mathrm{adpt}} \in \mathbb{R}^{L_P \times d_{\mathrm{in}}}$,
constructing
$X_{i,\mathrm{adpt}}A_k \in \mathbb{R}^{L_P \times r}$
for each agent $k=1,\ldots,N$ as these hidden states are produced.
These LR caches are therefore constructed from the hidden states generated by agent $i$ when $L_P$ is first processed, without separately processing $L_P$ for each agent.
As shown in Figure~\ref{fig:method_overview}(A1), the shared base cache and the LR caches for all agents cover $L_P$ when agent $i$ finishes its turn.
PreLRShared thereby replaces the later full-dimensional backbone processing over $L_P$ with lightweight rank-$r$ down-projections.
Agent $j$ subsequently reuses the shared base cache and its precomputed LR cache over $L_P$, as shown in Figure~\ref{fig:method_overview}(A2).
It therefore performs adapted backbone computation only for the newly added context $L_C$ rather than processing the accumulated trajectory again.
As agent $j$ processes $L_C$, PreLRShared applies every down-projection $A_k$ to the hidden states produced by agent $j$ and extends the corresponding LR caches.
After agent $j$ finishes its turn, the shared base cache and all LR caches cover $L_P + L_C$, as shown in Figure~\ref{fig:method_overview}(A3), supporting the same reuse by the next agent.

PreLRShared eliminates repeated backbone processing of the shared context, but its shared base cache remains constructed from the previous agent's adapter-conditioned hidden states.
We observe that this dependency can be reduced while retaining LR cache precomputation, motivating the adapter-free reconstruction introduced in ReBaseShared.

\subsection{ReBaseShared: Shared Base Cache Reconstruction}
\label{sec:method_rebase}

ReBaseShared reduces the dependence of the shared base cache on the previous agent by reconstructing it from adapter-free hidden states.
Although $W_0$ is shared across agents, $X_{i,\mathrm{adpt}}$ contains the effects of agent $i$'s adapters propagated from preceding layers.
Consequently, the base cache $X_{i,\mathrm{adpt}}W_0$ remains conditioned on the previous agent.
ReBaseShared instead processes the shared context with the LoRA adapters disabled.
We denote the resulting adapter-free hidden states by $X_{\mathrm{neut}}$ and construct the shared base cache as $X_{\mathrm{neut}}W_0$, which we refer to as the neutral base cache.

\begin{figure*}[t]
    \centering
    \vspace{-15pt}
    \includegraphics[width=1.0\linewidth]{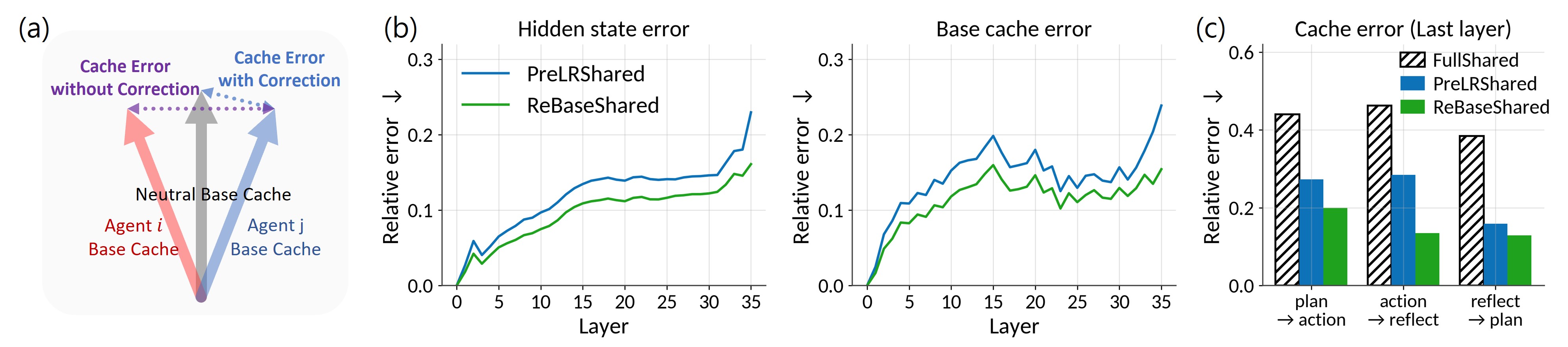}
    \vspace{-20pt}
    \caption{Effect of reconstructing the shared base cache from adapter-free hidden states in Ministral-8B on a HotpotQA trajectory.
    (a) compares the previous agent's and neutral base caches against the current agent's base cache.
    (b) reports the layer-wise relative errors in the hidden states and base caches of PreLRShared and ReBaseShared.
    (c) reports the last-layer relative error across agent transitions and includes FullShared as a reference for entire KV cache reuse.}
    \label{fig:neutral_error}
\end{figure*}

Figure~\ref{fig:neutral_error} examines whether this reconstruction brings the shared base cache closer to the base cache that the current agent would construct.
We use the base cache constructed from the current agent's adapted hidden states as the reference and report relative error against it.
Figure~\ref{fig:neutral_error}(a) illustrates this comparison, and Figure~\ref{fig:neutral_error}(b) shows that ReBaseShared reduces both the layer-wise hidden state error and the resulting base cache error relative to PreLRShared.
Figure~\ref{fig:neutral_error}(c) shows the same reduction at the last layer across all agent transitions in the trajectory.
FullShared exhibits a larger error because it reuses the entire KV cache constructed under the previous agent's adapter rather than sharing only the base component.
The experimental setup is shown in Section~\ref{sec:exp_setup}.
Appendix~\ref{app:neutral_derivation} derives the condition under which the neutral base cache has lower relative error, and Appendix~\ref{app:neutral_geometry} examines this condition empirically.

ReBaseShared retains LR cache precomputation, while constructing the other agents' LR caches from the same adapter-free hidden states used for neutral reconstruction.
As shown in Figure~\ref{fig:method_overview}(B1), before agent $j$ begins its turn, the neutral base cache and the precomputed LR cache for each agent already cover $L_P$.
For newly added context, agent $j$ uses its adapted hidden states $X_{j,\mathrm{adpt}}$ for generation and constructs its own LR cache $X_{j,\mathrm{adpt}}A_j$.
Meanwhile, adapter-free hidden states $X_{\mathrm{neut}}$ are used to construct the neutral base cache for subsequent reuse.
These adapter-free hidden states are also projected to $A_k$ for each other agent $k \neq j$ to construct its LR cache for subsequent use.

ReBaseShared$_{\mathrm{DB}}$ schedules the adapted and adapter-free paths together through \textbf{double-batching} during both prefill and decoding, as shown in Figure~\ref{fig:method_overview}(B2).
Both paths process the same tokens at the same positions while maintaining separate KV cache views.
The adapted path attends to the current agent's KV cache and produces the next-token logits.
In parallel, the adapter-free path processes the tokens, attending only to the neutral base cache and extending it for subsequent agents.
Thus, the neutral base cache constructed for $L_C$ does not affect agent $j$'s own generation.

ReBaseShared$_{\mathrm{LP}}$ uses \textbf{lazy-prefill}, executing only the adapted path during agent $j$'s turn, as shown in Figure~\ref{fig:method_overview}(B3).
After agent $j$ finishes decoding, it processes $L_C$ once with the adapters disabled, as shown in Figure~\ref{fig:method_overview}(B4).
This contiguous prefill starts from the neutral base cache over $L_P$ and extends it over $L_C$ before the next agent begins its turn.

After either schedule completes, the neutral base cache and the LR caches cover the full trajectory $L_P + L_C$, as shown in Figure~\ref{fig:method_overview}(B5).
In both schedules, the adapter-free path processes the same token sequence at the same positions while attending to the same neutral base cache.
DB and LP therefore produce logically equivalent cache states and differ only in when the adapter-free reconstruction is performed.
LP performs the reconstruction as a contiguous prefill after the current turn and avoids maintaining an adapter-free decoding path, making it suitable for single-stream edge inference.
DB incorporates the adapted and adapter-free paths into the continuous serving batch during prefill and decoding, making it suitable for concurrent server-side serving~\citep{kwon2023vllm,zheng2024sglang}.
Under concurrent serving, this schedule reduces the queueing delay caused by executing reconstruction as a separate prefill~\citep{agrawal2024sarathi,zhong2024distserve}.

\section{Experiments}
\label{sec:exp}

\subsection{Implementation Setup}
\label{sec:exp_setup}

\textbf{Agent Setup.}
We use the multi-hop agent framework of AutoAct~\citep{qiao2024autoact}, consisting of three role-specific agents for planning, action, and reflection.
The plan and action agents alternate to obtain observations, after which the reflect agent either begins another cycle or returns the final answer.
The agents access web search through the Serper API~\citep{serper} and Wikipedia lookup~\citep{yao2023react}.

\textbf{Models and Datasets.}
We use LLaMA-3.1-8B-Instruct~\citep{grattafiori2024llama3} and Ministral-8B-Instruct~\citep{mistral2024ministral} on HotpotQA~\citep{yang2018hotpotqa} and ScienceQA~\citep{lu2022scienceqa}.
Following AutoAct~\citep{qiao2024autoact}, we consider three difficulty levels for each benchmark.

\textbf{Training Settings.}
We train a separate LoRA adapter for each role using the corresponding synthetic and filtered AutoAct trajectories.
LoRA is applied to the query and value projections with rank $r=8$, while the remaining training settings follow LRAgent~\citep{jeon2026lragent}.
Under this QV setting, the key projection has no LR component and is fully shared, while the value cache is decomposed into base and LR caches.
We provide ablations on the LoRA rank in Appendix~\ref{app:lora_rank}, showing that accuracy saturates from $r=8$ and supporting its use as the default rank.
We further evaluate LoRA applied to all attention projections in Appendix~\ref{app:lora_qkvo}, showing that PReCache retains its efficiency benefits beyond the default QV setting.

\textbf{Baselines.}
We compare against NonShared, FullShared, DroidSpeak~\citep{liu2026droidspeak}, CacheBlend~\citep{yao2025cacheblend}, RelayCaching~\citep{geng2026relaycaching}, and BaseShared~\citep{jeon2026lragent}.
The selective recomputation methods use a $30\%$ recomputation ratio, which lies near the accuracy-efficiency Pareto frontier for DroidSpeak and provides a practical operating point with sufficient accuracy recovery for CacheBlend and RelayCaching~\citep{jeon2026lragent}.
BaseLRShared is excluded because it requires adapters trained with a shared LoRA down-projection and therefore does not support existing adapters with different down-projection weights.
Under our matched-prefix setting, the original correction variables of prefix-based correction methods~\citep{ye2025kvcomm,li2026graphflow,ma2026kamera} remain unchanged and provide no adapter-specific correction.
Their application is therefore equivalent to FullShared for differences introduced by LoRA adapters.

\textbf{Efficiency Setup.}
To isolate the overhead of each KV cache sharing scheme from variations in tool latency and generation, we replay controlled traces with identical agent schedules and token counts across methods.
We vary the retrieved context to construct three-agent trajectories ranging from $1.9$k to $66.4$k tokens.
For single-stream inference, TTFT is summed across agent calls.
Per-request throughput is computed from the trajectory length and accumulated agent-call latency.
For single-stream edge inference, we run one trajectory at a time on a single NVIDIA A6000 48GB GPU using ReBaseShared$_{\mathrm{LP}}$.
For concurrent server-side serving, we use vLLM~\citep{kwon2023vllm} on a single NVIDIA A100 80GB GPU with chunked prefill, paged KV cache management, and prefix caching enabled.
We sweep the request rate from $0.5$ to $16$ queries per second (QPS) using a fixed $17.3$k-token trajectory and ReBaseShared$_{\mathrm{DB}}$.
Methods using LR caches employ Flash-LoRA-Attention~\citep{jeon2026lragent}.

\subsection{Benchmark Accuracy Evaluation}
\label{sec:exp_acc}

\begin{table*}[t]
\centering
\vspace{-10pt}
\caption{Mean benchmark accuracy (\%) of NonShared and KV cache sharing methods on HotpotQA and ScienceQA.
The value beside each average denotes its percentage-point difference from NonShared.
For each backbone and benchmark, the higher and lower halves of the methods ranked by average accuracy are highlighted in green and red, respectively.}
\label{tab:accuracy}
\resizebox{\linewidth}{!}{
\begin{tabular}{llcccccccc}
\toprule
& & \multicolumn{4}{c}{\textbf{HotpotQA}} & \multicolumn{4}{c}{\textbf{ScienceQA}} \\
\cmidrule(lr){3-6} \cmidrule(lr){7-10}
\textbf{Model} & \textbf{Method}
& Easy & Medium & Hard & \textbf{Avg.}
& 1--4 & 5--8 & 9--12 & \textbf{Avg.} \\
\midrule
\multirow{8}{*}{\footnotesize LLaMA-3.1-8B}
& \footnotesize NonShared
& 40.55 & 41.05 & 30.95 & 37.52\ {\tiny 0.00}
& 70.33 & 59.67 & 77.58 & 69.19\ {\tiny 0.00} \\
& \footnotesize FullShared
& 38.05 & 37.90 & 26.15 & 34.03\ {\tiny $-$3.48}
& 67.71 & 56.42 & 72.75 & 65.62\ {\tiny $-$3.57} \\
& \footnotesize DroidSpeak
& 39.80 & 38.30 & 27.60 & 35.23\ {\tiny \textcolor{CustomRed}{$-$2.28}}
& 68.17 & 59.04 & 74.92 & 67.37\ {\tiny \textcolor{CustomRed}{$-$1.82}} \\
& \footnotesize CacheBlend
& 39.15 & 38.55 & 27.10 & 34.93\ {\tiny \textcolor{CustomRed}{$-$2.59}}
& 67.90 & 58.38 & 73.71 & 66.66\ {\tiny \textcolor{CustomRed}{$-$2.53}} \\
& \footnotesize RelayCaching
& 40.15 & 39.55 & 29.20 & 36.30\ {\tiny \textcolor{CustomRed}{$-$1.22}}
& 68.79 & 59.25 & 75.83 & 67.96\ {\tiny \textcolor{CustomRed}{$-$1.23}} \\
& \footnotesize BaseShared
& 40.40 & 40.35 & 30.15 & 36.97\ {\tiny \textcolor{CustomGreen}{$-$0.55}}
& 69.54 & 59.50 & 76.88 & 68.64\ {\tiny \textcolor{CustomGreen}{$-$0.56}} \\
& \footnotesize PreLRShared
& 40.35 & 40.35 & 28.25 & 36.32\ {\tiny \textcolor{CustomGreen}{$-$1.20}}
& 69.46 & 58.42 & 76.79 & 68.22\ {\tiny \textcolor{CustomGreen}{$-$0.97}} \\
& \footnotesize ReBaseShared
& 40.65 & 39.95 & 30.25 & 36.95\ {\tiny \textcolor{CustomGreen}{$-$0.57}}
& 69.74 & 59.41 & 77.17 & 68.77\ {\tiny \textcolor{CustomGreen}{$-$0.42}} \\
\midrule
\multirow{8}{*}{\footnotesize Ministral-8B}
& \footnotesize NonShared
& 41.95 & 38.35 & 29.85 & 36.72\ {\tiny 0.00}
& 70.96 & 65.13 & 71.13 & 69.07\ {\tiny 0.00} \\
& \footnotesize FullShared
& 37.40 & 36.10 & 26.60 & 33.37\ {\tiny $-$3.35}
& 69.96 & 57.88 & 62.33 & 63.39\ {\tiny $-$5.68} \\
& \footnotesize DroidSpeak
& 39.05 & 36.85 & 28.20 & 34.70\ {\tiny \textcolor{CustomRed}{$-$2.02}}
& 70.00 & 61.04 & 66.25 & 65.76\ {\tiny \textcolor{CustomRed}{$-$3.31}} \\
& \footnotesize CacheBlend
& 38.20 & 36.55 & 27.25 & 34.00\ {\tiny \textcolor{CustomRed}{$-$2.72}}
& 69.54 & 59.71 & 64.46 & 64.57\ {\tiny \textcolor{CustomRed}{$-$4.50}} \\
& \footnotesize RelayCaching
& 39.20 & 36.75 & 27.95 & 34.63\ {\tiny \textcolor{CustomRed}{$-$2.09}}
& 69.50 & 61.00 & 65.88 & 65.46\ {\tiny \textcolor{CustomRed}{$-$3.61}} \\
& \footnotesize BaseShared
& 40.75 & 37.25 & 28.30 & 35.43\ {\tiny \textcolor{CustomGreen}{$-$1.28}}
& 69.13 & 62.42 & 68.67 & 66.74\ {\tiny \textcolor{CustomGreen}{$-$2.33}} \\
& \footnotesize PreLRShared
& 39.15 & 37.05 & 28.10 & 34.77\ {\tiny \textcolor{CustomGreen}{$-$1.95}}
& 68.04 & 61.46 & 67.83 & 65.78\ {\tiny \textcolor{CustomGreen}{$-$3.29}} \\
& \footnotesize ReBaseShared
& 40.70 & 37.85 & 29.70 & 36.08\ {\tiny \textcolor{CustomGreen}{$-$0.62}}
& 64.71 & 64.75 & 69.96 & 66.47\ {\tiny \textcolor{CustomGreen}{$-$2.60}} \\
\bottomrule
\end{tabular}
}
\vspace{-10pt}
\end{table*}

Table~\ref{tab:accuracy} reports the benchmark accuracy of all methods.
Across the four backbone and benchmark combinations, ReBaseShared remains closest to NonShared overall, with an average accuracy drop of $1.1\%$ points and accuracy comparable to BaseShared.
PreLRShared exhibits average degradation ranging from $1.0\%$ to $3.3\%$ points, while ReBaseShared reduces this degradation on average by reconstructing the shared base cache from adapter-free hidden states.
ReBaseShared achieves accuracy comparable to BaseShared while eliminating most of its repeated prefill computation.
PreLRShared also remains comparable to or more accurate than the selective recomputation baselines, demonstrating the benefit of retaining an agent-specific LR cache even without neutral base cache reconstruction.

Appendix~\ref{app:acc_latency} reports latency and trajectory length under naturally generated agent trajectories, showing that PreLRShared and ReBaseShared remain in a favorable accuracy-latency region even when cache sharing changes trajectory behavior.
Appendix~\ref{app:acc_dev} reports standard deviations across 20 complete benchmark runs, showing low run-to-run variation.

\subsection{System Efficiency}
\label{sec:exp_eff}

\begin{figure*}[t]
    \centering
    \vspace{-10pt}
    \includegraphics[width=1.0\linewidth]{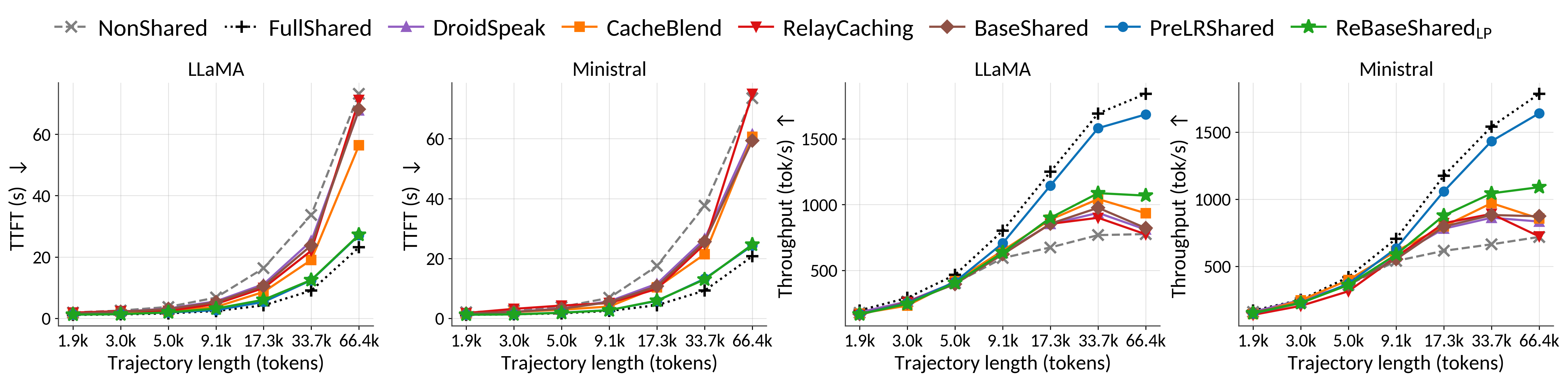}
    \vspace{-20pt}
    \caption{Single-stream TTFT ($\downarrow$) and per-request throughput ($\uparrow$) over trajectory length for both backbones on a single A6000 GPU. ReBaseShared uses lazy prefill scheduling.}
    \label{fig:efficiency}
    \vspace{-5pt}
\end{figure*}

\textbf{Single-Stream Efficiency.}
Figure~\ref{fig:efficiency} shows that the efficiency advantages of PreLRShared and ReBaseShared$_{\mathrm{LP}}$ increase as the shared trajectory grows.
By replacing repeated full-length backbone processing with rank-$r$ down-projections, PreLRShared closely approaches FullShared and remains within $17\%$ of its TTFT at $66.4$k tokens.
Across both backbones, PreLRShared reduces TTFT by up to $3.1\times$ and improves per-request throughput by up to $2.3\times$ over NonShared.

ReBaseShared$_{\mathrm{LP}}$ reconstructs the neutral base cache through one adapter-free pass over each newly added context.
Because this reconstruction occurs after decoding, ReBaseShared$_{\mathrm{LP}}$ closely follows PreLRShared in TTFT, while its additional end-to-end computation results in lower throughput.
Nevertheless, it reduces TTFT by up to $3.0\times$ and improves per-request throughput by up to $1.6\times$ over NonShared, while outperforming all selective recomputation baselines at long trajectories.

BaseShared remains close to NonShared and DroidSpeak because it still requires agent-specific backbone processing over the accumulated context.
CacheBlend and RelayCaching reduce this processing through token-wise recomputation, but increasing their accuracy requires full-dimensional KV cache computation for more tokens.
RelayCaching additionally computes attention scores to select tokens during inference.
In contrast, PreLRShared limits additional cache construction to rank-$r$ down-projections, while ReBaseShared retains this precomputation with one adapter-free reconstruction.
Appendix~\ref{app:eff_edge} provides the numerical results for all trajectory lengths.

\begin{figure*}[t]
    \centering
    \includegraphics[width=1.0\linewidth]{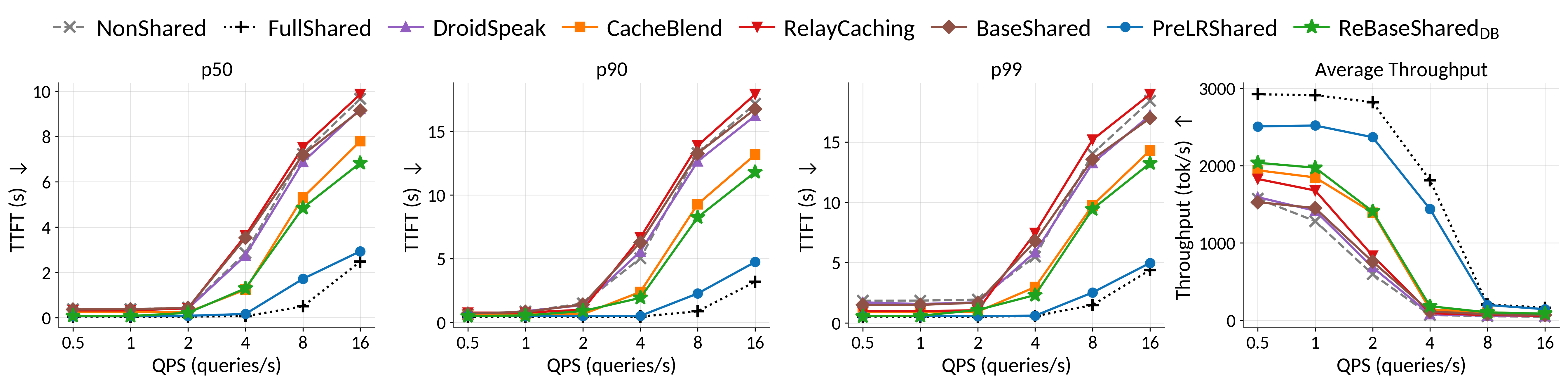}
    \vspace{-20pt}
    \caption{Serving TTFT percentiles ($\downarrow$) and per-request throughput ($\uparrow$) over request rate on vLLM with a fixed $17.3$k-token trajectory on a single A100 GPU. ReBaseShared uses double batching.}
    \label{fig:vllm}
    \vspace{-5pt}
\end{figure*}

\textbf{Concurrent Serving.}
Figure~\ref{fig:vllm} shows that PreLRShared and ReBaseShared$_{\mathrm{DB}}$ retain their efficiency advantages over BaseShared and the selective recomputation baselines under concurrent serving at high loads.
At low request rates, when queueing remains limited, KV cache sharing methods exhibit similar TTFT.
As the load increases, methods with greater computation saturate earlier, whereas PreLRShared and ReBaseShared$_{\mathrm{DB}}$ sustain lower TTFT over a wider request-rate range.
The widening gaps at high request rates therefore reflect differences in serving capacity.
Under unsaturated load, PreLRShared and ReBaseShared provide $1.6\times$ and $1.3\times$ higher per-request throughput than NonShared, respectively.

DB incorporates the adapted and adapter-free paths into the continuous serving batch during prefill and decoding.
By distributing reconstruction throughout the current turn, DB avoids a separate post-turn prefill that can delay queued requests.
The p90 and p99 results exhibit the same saturation trend, with larger gaps as queueing accumulates in the tail.
Appendix~\ref{app:eff_server} provides the numerical results across all request rates and TTFT percentiles.

\begin{figure*}[t]
    \centering
    \vspace{-10pt}
    \includegraphics[width=1.0\linewidth]{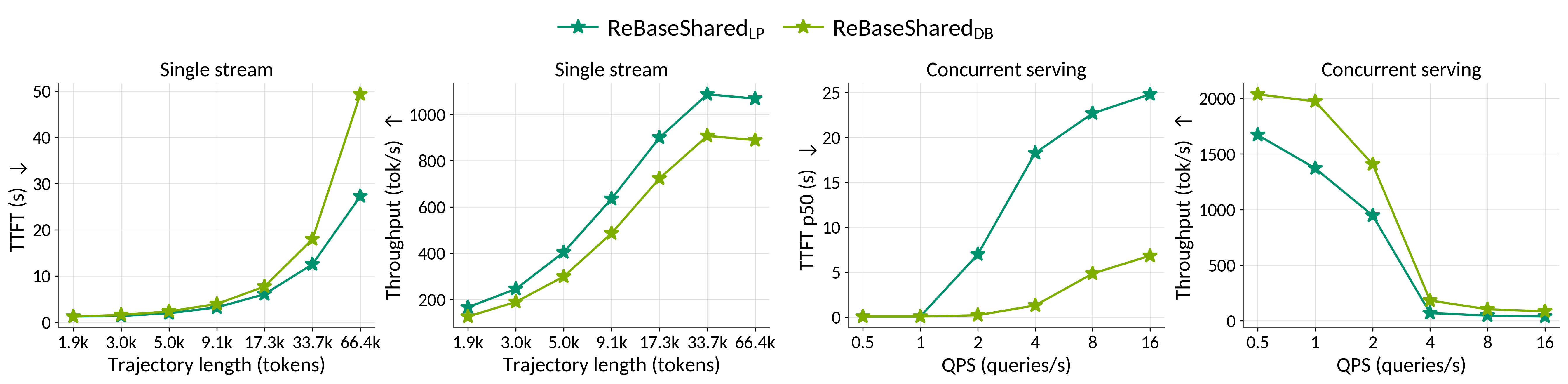}
    \vspace{-15pt}
    \caption{Efficiency comparison of LP and DB scheduling for ReBaseShared. The left panels report single-stream TTFT ($\downarrow$) and per-request throughput ($\uparrow$) over trajectory length. The right panels report p50 TTFT ($\downarrow$) and per-request throughput ($\uparrow$) over request rate with a fixed $17.3$k-token trajectory under concurrent serving.}
    \label{fig:schedule}
    \vspace{-10pt}
\end{figure*}

\textbf{LP vs. DB Scheduling.}
Figure~\ref{fig:schedule} shows how LP and DB support efficient neutral base cache reconstruction in single-stream inference and concurrent serving, respectively.
In single-stream inference, LP performs adapter-free reconstruction as a contiguous prefill after the current agent finishes decoding, whereas DB executes the adapted and adapter-free paths together during both prefill and decoding.
LP therefore keeps reconstruction outside the next agent's TTFT and avoids maintaining an adapter-free decoding path, resulting in lower TTFT and higher throughput than DB.

Under concurrent serving, DB incorporates the adapter-free path into the continuous serving batch throughout the current turn.
This avoids delaying queued requests with a separate reconstruction prefill after each turn, allowing DB to sustain lower TTFT and higher throughput than LP as the request rate increases.
Appendix~\ref{app:eff_schedule} reports the numerical results, confirming the respective advantages of LP and DB in the single-stream and concurrent settings.

\begin{wrapfigure}{l}{0.42\linewidth}
    \centering
    \vspace{-5pt}
    \includegraphics[width=\linewidth]{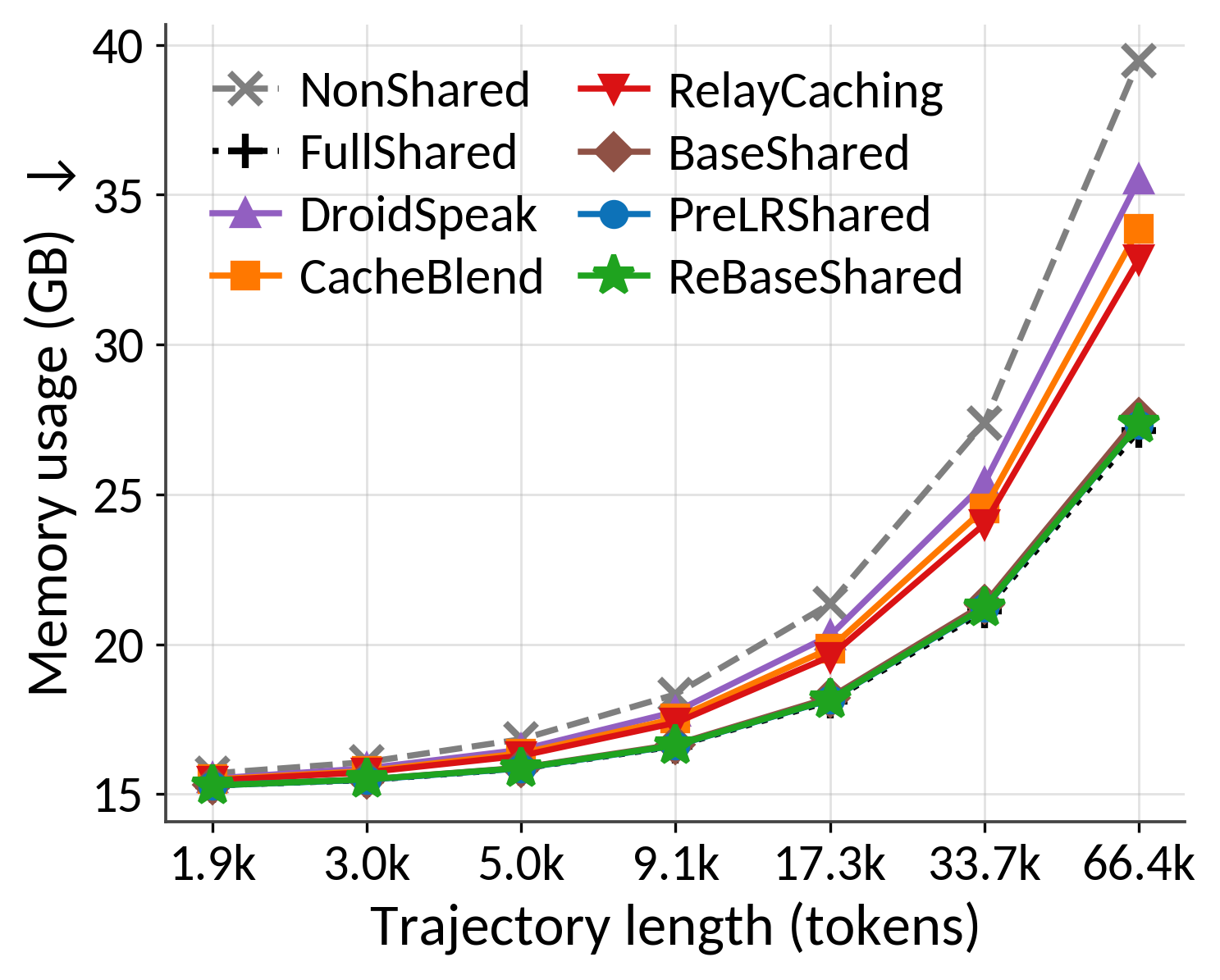}
    \vspace{-20pt}
    \caption{Peak GPU memory usage across trajectory lengths for LLaMA-3.1-8B.}
    \label{fig:memory}
    \vspace{-5pt}
\end{wrapfigure}

\textbf{Memory Usage.}
Figure~\ref{fig:memory} shows that PreLRShared and ReBaseShared retain the memory efficiency of BaseShared, remaining close to FullShared and below all selective recomputation baselines as the trajectory grows.
Both methods store one shared base cache and compact per-agent LR caches, avoiding full-dimensional KV cache replication across agents.
Precomputing the LR caches for all agents adds little memory because each cache has rank $r$, whereas selective recomputation retains full-dimensional agent-specific KV caches for the recomputed layers or tokens.
At $66.4$k tokens, PreLRShared and ReBaseShared remain within $2\%$ of FullShared while using $17$--$23\%$ less peak memory than the selective recomputation baselines.
DroidSpeak incurs the largest overhead by additionally retaining hidden states before the first recomputed layer, while CacheBlend uses more memory than RelayCaching because it recomputes selected tokens across all layers.
Appendix~\ref{app:eff_memory} provides the complete measurements, and Appendix~\ref{app:eff_agentn} shows that this memory efficiency is maintained as the number of agents increases.

\section{Conclusion}

In this work, we present PReCache, a training-free KV cache sharing framework comprising two complementary designs, PreLRShared and ReBaseShared, for existing multi-LoRA agents.
PreLRShared eliminates repeated processing of the accumulated trajectory by precomputing compact agent-specific LR caches when each context segment is first processed.
ReBaseShared further reduces the previous-agent dependency of the shared base cache by reconstructing it from adapter-free hidden states while retaining LR cache precomputation.
We further develop LP and DB as inference schedules tailored to single-stream edge inference and concurrent server-side serving, respectively.
Across multiple models and benchmarks, ReBaseShared maintains accuracy within an average of $1.1\%$ points of NonShared, comparable to BaseShared, while avoiding most of the repeated prefill computation retained by BaseShared.
PreLRShared achieves up to a $3.1\times$ TTFT speedup and $2.3\times$ higher per-request throughput than NonShared in single-stream inference.
Both designs maintain peak memory within $2\%$ of FullShared at the longest trajectory and retain their efficiency advantages under concurrent serving.
Overall, PReCache reduces KV cache memory usage and repeated prefill computation while preserving agent-specific behavior, without retraining or restricting existing LoRA adapters.

\newpage
\bibliography{iclr2027_conference}
\bibliographystyle{iclr2027_conference}

\newpage
\appendix

\section{Analysis of Neutral Base Cache Reconstruction}
\label{app:neutral_analysis}

\subsection{Derivation}
\label{app:neutral_derivation}

As described in Section~\ref{sec:method_rebase}, ReBaseShared replaces the base cache constructed from the previous agent's adapted hidden states with a neutral base cache constructed from adapter-free hidden states.
We derive the condition under which the neutral base cache has lower relative error against the base cache constructed directly by the current agent.

Consider a fixed layer and token position where agent $j$ receives context previously processed by agent $i$.
We denote the base caches constructed from the previous agent's, current agent's, and adapter-free hidden states by
\begin{equation}
    C_i = X_{i,\mathrm{adpt}}W_0,\qquad
    C_j = X_{j,\mathrm{adpt}}W_0,\qquad
    C_{\mathrm{neut}} = X_{\mathrm{neut}}W_0.
\end{equation}
Their offsets from the neutral base cache are
\begin{equation}
    \Delta_i = C_i-C_{\mathrm{neut}},\qquad
    \Delta_j = C_j-C_{\mathrm{neut}}.
\end{equation}
All norms below are Euclidean norms over the cache vector at the fixed layer and token position.

Using the previous agent's base cache gives the relative error
\begin{equation}
    E_{\mathrm{prev}}
    = \frac{\lVert C_i-C_j\rVert_2}{\lVert C_j\rVert_2}
    = \frac{\lVert\Delta_i-\Delta_j\rVert_2}{\lVert C_j\rVert_2},
\end{equation}
whereas using the neutral base cache gives
\begin{equation}
    E_{\mathrm{neut}}
    = \frac{\lVert C_{\mathrm{neut}}-C_j\rVert_2}{\lVert C_j\rVert_2}
    = \frac{\lVert\Delta_j\rVert_2}{\lVert C_j\rVert_2}.
\end{equation}

For nonzero $\Delta_i$ and $\Delta_j$, we define their norm ratio and cosine similarity as
\begin{equation}
    \rho
    = \frac{\lVert\Delta_i\rVert_2}{\lVert\Delta_j\rVert_2},
    \qquad
    \cos\theta
    = \frac{\langle\Delta_i,\Delta_j\rangle}
    {\lVert\Delta_i\rVert_2\lVert\Delta_j\rVert_2}.
\end{equation}
The ratio between the relative errors is then
\begin{equation}
    \Gamma_{\mathrm{err}}
    = \frac{E_{\mathrm{prev}}}{E_{\mathrm{neut}}}
    = \sqrt{1+\rho^2-2\rho\cos\theta}.
\end{equation}
Therefore,
\begin{equation}
    E_{\mathrm{prev}} \ge E_{\mathrm{neut}}
    \quad\Longleftrightarrow\quad
    \Gamma_{\mathrm{err}} \ge 1
    \quad\Longleftrightarrow\quad
    \rho \ge 2\cos\theta.
    \label{eq:neutral-closer}
\end{equation}

Thus, $\Gamma_{\mathrm{err}}>1$ indicates that the neutral base cache has lower relative error than the base cache constructed from the previous agent's hidden states.
When $\rho = 2\cos\theta$, the two base caches have equal relative error.
The same derivation applies to hidden states by replacing $C_i$, $C_j$, and $C_{\mathrm{neut}}$ with their corresponding hidden states.
Therefore, the condition in Equation~\ref{eq:neutral-closer} characterizes when the adapter-free reconstruction has no greater error against the current agent's state, with a strict reduction when the inequality holds strictly.

\newpage
\subsection{Measured Geometry}
\label{app:neutral_geometry}

\begin{figure}[h]
    \centering
    \includegraphics[width=1.0\linewidth]{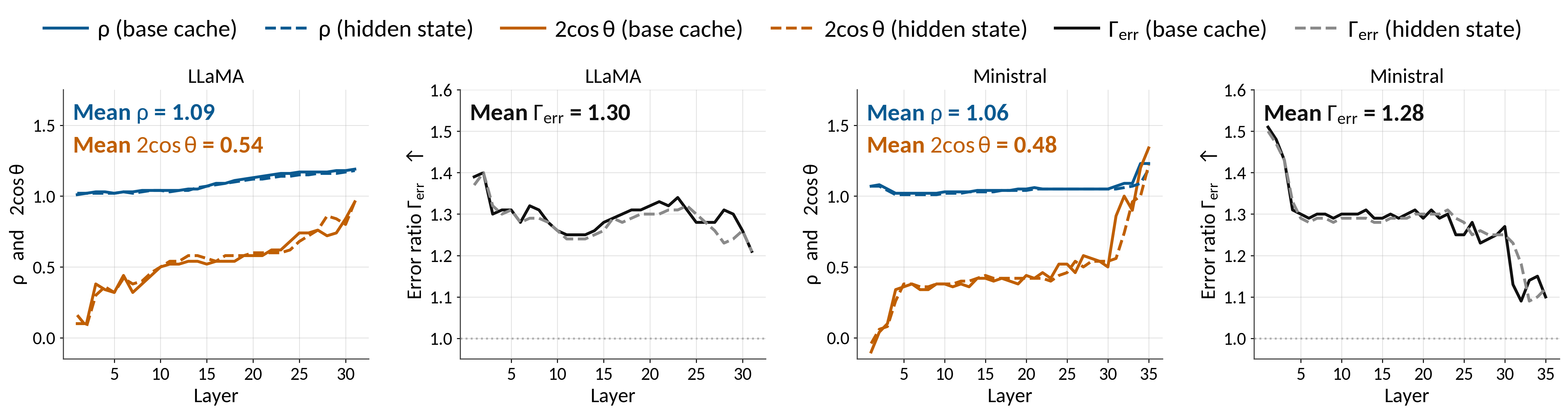}
    \caption{Layer-wise geometry of neutral base cache reconstruction in ReBaseShared.
    Each quantity is computed per token and then averaged across tokens, prompts, and agent transitions.
    Solid and dashed lines represent base cache and hidden state measurements, respectively.
    At the token level, $\rho>2\cos\theta$, or equivalently $\Gamma_{\mathrm{err}}>1$, indicates lower relative error for the neutral state than for the state constructed from the previous agent's hidden states.}
    \label{fig:neutral_geometry}
\end{figure}

Appendix~\ref{app:neutral_derivation} shows that the neutral base cache has lower relative error than the base cache constructed from the previous agent's hidden states when $\rho > 2\cos\theta$, or equivalently when $\Gamma_{\mathrm{err}}>1$.
We measure these quantities for LLaMA-3.1-8B-Instruct and Ministral-8B-Instruct on the three agent transitions in HotpotQA: plan-to-action, action-to-reflect, and reflect-to-plan.
For each layer, we first compute $\rho$, $2\cos\theta$, and $\Gamma_{\mathrm{err}}$ separately for each token and then average them across tokens, prompts, and agent transitions.
Here, $\rho$ represents the ratio between the magnitudes of the previous and current agents' offsets from the neutral state, while $2\cos\theta$ gives the boundary derived in Equation~\ref{eq:neutral-closer}.
Figure~\ref{fig:neutral_geometry} reports the resulting layer-wise averages for both hidden states and base caches.

As shown in Figure~\ref{fig:neutral_geometry}, the layer-wise average of $\Gamma_{\mathrm{err}}$ remains above one at every measured layer for both backbones and for both hidden states and base caches.
Thus, the token-wise error ratio favors the neutral representation on average at each layer.
The averaged $\rho$ and $2\cos\theta$ exhibit a consistent trend, with the average $\rho$ exceeding the average boundary over most layers.

This analysis isolates the change from the base cache used by PreLRShared to the neutral base cache used by ReBaseShared.
BaseShared uses the same previous-agent base cache as PreLRShared, so base cache geometry alone does not capture its primary accuracy benefit.
Instead, BaseShared processes the shared context with the current agent's adapter to compute its LR cache.
ReBaseShared reduces the base cache error without this repeated backbone processing, yielding accuracy comparable to BaseShared through a different KV cache construction, as shown in Section~\ref{sec:exp_acc}.

Overall, these measurements support the error reduction predicted in Appendix~\ref{app:neutral_derivation} and observed in Figure~\ref{fig:neutral_error}.
Across the measured backbones and agent transitions, the neutral base cache remains closer on average to the base cache constructed directly by the current agent than the base cache constructed from the previous agent's hidden states.
ReBaseShared reduces the average degradation of PreLRShared across all four backbone--benchmark pairs, although the improvement varies across individual difficulty groups.
Our main evaluation uses HotpotQA and ScienceQA, for which role-specific multi-LoRA training trajectories are available.
Prior analysis across additional agent-role structures and task domains also shows that the base cache remains more similar across agents than the entire KV cache~\citep{jeon2026lragent}, providing additional evidence that the shared base decomposition is not specific to the plan--action--reflect workflow.
We note that this analysis focuses on the shared base cache, which is directly reused across agents and is the primary target of neutral reconstruction.
The agent-specific LR caches are maintained separately, and their differences remain small after the corresponding agent-specific down-projections.

\newpage

\section{Accuracy Benchmark Analysis}
\label{app:acc}

\subsection{Benchmark Latency}
\label{app:acc_latency}

\begin{figure*}[h]
    \centering
    \includegraphics[width=0.95\linewidth]{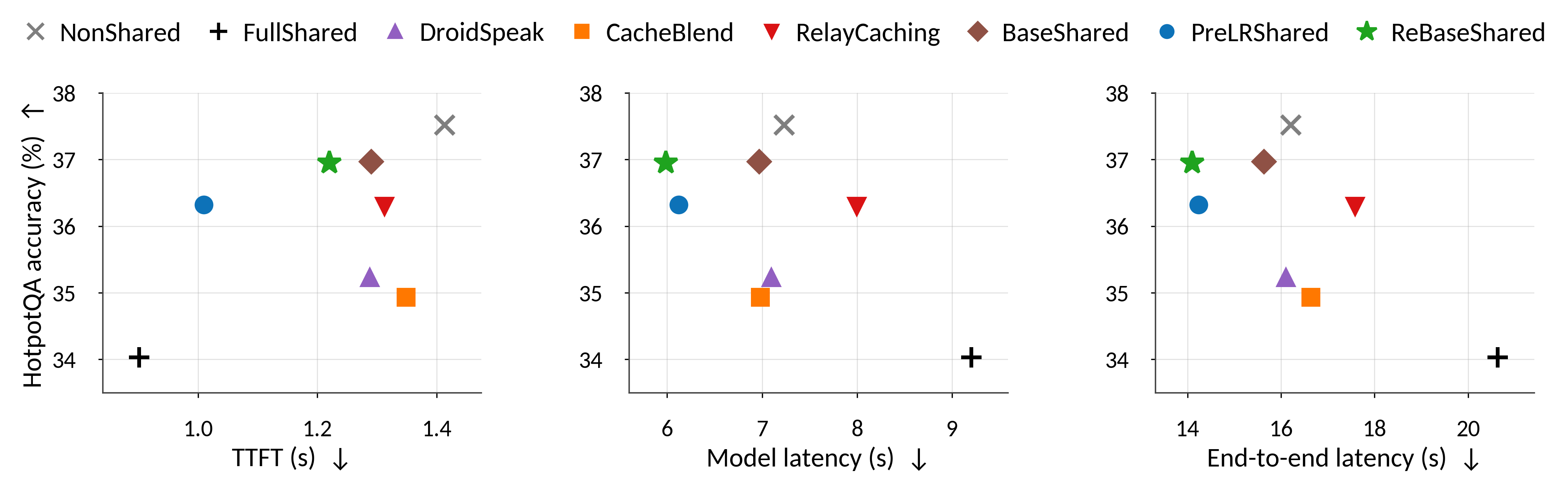}
    \caption{Accuracy-latency trade-off for LLaMA-3.1-8B on HotpotQA. The panels compare benchmark accuracy with TTFT, model latency, and end-to-end latency, respectively. Higher accuracy and lower latency are preferred.}
    \label{fig:bench_latency}
\end{figure*}

As discussed in Section~\ref{sec:exp_acc}, KV cache sharing changes generated tokens and subsequent agent decisions, altering the number of agent steps and the resulting trajectory length.
While the main efficiency evaluation uses controlled traces to isolate the computation of each sharing method, this appendix measures latency on actual benchmark trajectories to capture the combined effects of cache sharing on model execution and agent behavior.
TTFT is measured for each agent invocation and summed across the complete trajectory.
Model latency includes all model execution, including prefill, decoding, and cache reconstruction when applicable, while end-to-end latency additionally includes tool calls and other agent-side overhead.
For ReBaseShared$_{\mathrm{LP}}$, post-turn reconstruction is included in model and end-to-end latency but not in the TTFT of the next agent invocation.
Table~\ref{tab:bench_latency} reports these metrics and trajectory lengths for LLaMA-3.1-8B on HotpotQA, and Figure~\ref{fig:bench_latency} summarizes the resulting accuracy-latency trade-off.

\begin{table*}[h]
\centering
\caption{Average benchmark latency and trajectory length for LLaMA-3.1-8B on HotpotQA. Latencies are reported in seconds, with average and maximum trajectory lengths reported in tokens.}
\label{tab:bench_latency}
\resizebox{\linewidth}{!}{
\begin{tabular}{lccccc}
\toprule
\multirow{2}{*}{Method}
& \multirow{2}{*}{TTFT (s)}
& \multirow{2}{*}{Model Latency (s)}
& \multirow{2}{*}{E2E Latency (s)}
& \multicolumn{2}{c}{Trajectory Length (tokens)} \\
\cmidrule(lr){5-6}
&
&
&
&
Average
& Maximum \\
\midrule
NonShared
& 1.41 & 7.23 & 16.21 & 1268 & 6857 \\
FullShared
& 0.90 & 9.20 & 20.64 & 1757 & 7078 \\
DroidSpeak
& 1.29 & 7.10 & 16.11 & 1340 & 6317 \\
CacheBlend
& 1.35 & 6.98 & 16.64 & 1569 & 7103 \\
RelayCaching
& 1.31 & 8.00 & 17.58 & 1455 & 6982 \\
BaseShared
& 1.29 & 6.97 & 15.64 & 1224 & 6553 \\
PreLRShared
& 1.01 & 6.12 & 14.24 & 1416 & 5980 \\
ReBaseShared$_{\mathrm{LP}}$
& 1.22 & 5.99 & 14.10 & 1240 & 7884 \\
\bottomrule
\end{tabular}
}
\end{table*}

PreLRShared and ReBaseShared$_{\mathrm{LP}}$ provide the most favorable overall accuracy-latency trade-offs, remaining closest to the upper-left region of Figure~\ref{fig:bench_latency}.
Methods with greater accuracy degradation generally require more agent steps and produce longer trajectories, increasing model and end-to-end latency even when they reduce TTFT.
The gap between the average and maximum trajectory lengths further shows that some requests grow substantially longer than the average, reinforcing the importance of memory-efficient KV cache sharing for long-horizon trajectories.

\newpage
\subsection{Accuracy Deviation}
\label{app:acc_dev}

While Section~\ref{sec:exp_acc} reports mean benchmark accuracy, this appendix examines its variation across repeated evaluations.
For each method and benchmark setting, we repeat the complete benchmark evaluation 20 times and compute the standard deviation across these runs.
Table~\ref{tab:acc_dev} reports the resulting standard deviations for HotpotQA and ScienceQA.

\begin{table*}[h]
\centering
\caption{Standard deviation of benchmark accuracy across 20 complete evaluation runs, reported in percentage points.}
\label{tab:acc_dev}
\resizebox{\linewidth}{!}{
\begin{tabular}{lrrrrrrrr}
\toprule
Benchmark
& NonShared
& FullShared
& DroidSpeak
& CacheBlend
& RelayCaching
& BaseShared
& PreLRShared
& ReBaseShared \\
\midrule
HotpotQA
& 0.21 & 0.44 & 0.28 & 0.34 & 0.31 & 0.25 & 0.30 & 0.32 \\
ScienceQA
& 0.28 & 0.60 & 0.34 & 0.35 & 0.39 & 0.35 & 0.39 & 0.37 \\
\bottomrule
\end{tabular}
}
\end{table*}

The standard deviations remain below $0.60\%$ points for all methods.
Thus, the accuracy degradation of FullShared and selective recomputation generally exceeds the observed run-to-run variation, whereas the differences between BaseShared and ReBaseShared remain small.
Variations in external search results and model generation contribute to the remaining differences across runs.

\newpage
\section{Throughput, TTFT, and Memory Usage}
\label{app:eff}

\subsection{Single-Stream Edge Inference}
\label{app:eff_edge}

Section~\ref{sec:exp_eff} presents the single-stream efficiency trends over trajectory length, while Table~\ref{tab:eff_edge} provides the complete numerical results.
The experiments use controlled three-agent trajectories ranging from $1.9$k to $66.4$k tokens on LLaMA-3.1-8B and Ministral-8B with a single A6000 GPU.
PreLRShared and ReBaseShared$_{\mathrm{LP}}$ exhibit similar TTFT because LP reconstruction remains outside the TTFT of the next agent invocation.
At longer trajectories, PreLRShared achieves higher throughput because ReBaseShared$_{\mathrm{LP}}$ includes an additional adapter-free reconstruction in its inference time.
For single-stream inference, TTFT is measured from the start of each agent invocation to its first output token and summed across the trajectory.
Per-request throughput is the trajectory length divided by the total execution time, including LP reconstruction.
For ReBaseShared$_{\mathrm{LP}}$, reconstruction is included in this completion time but occurs before the next agent invocation and is therefore excluded from its TTFT.
While the accuracy evaluation is conducted on naturally generated benchmark trajectories, these controlled traces extend to longer trajectories to characterize system scaling beyond the lengths covered by the benchmarks.
These traces retain the short decoding segments of the agent trajectories, reflecting their prefill-dominated execution.

\begin{table*}[h]
\centering
\caption{Single-stream TTFT in seconds and throughput in tokens per second over trajectory length on a single A6000 GPU. ReBaseShared uses LP scheduling, and TP denotes throughput.}
\label{tab:eff_edge}
\resizebox{\linewidth}{!}{
\begin{tabular}{lllrrrrrrr}
\toprule
Model & Metric & Method
& 1.9k & 3.0k & 5.0k & 9.1k & 17.3k & 33.7k & 66.4k \\
\midrule

\multirow{16}{*}{LLaMA-3.1-8B}
& \multirow{8}{*}{TTFT (s)}
& NonShared     & 1.93 & 2.53 & 3.72 & 6.77 & 16.36 & 33.72 & 73.18 \\
&
& FullShared    & 1.14 & 1.27 & 1.63 & 2.39 & 4.22 & 9.05 & 23.15 \\
&
& DroidSpeak    & 1.62 & 2.15 & 3.21 & 5.55 & 11.12 & 25.23 & 67.82 \\
&
& CacheBlend    & 1.67 & 2.15 & 2.31 & 3.72 & 8.61 & 19.05 & 56.38 \\
&
& RelayCaching  & 1.87 & 2.35 & 2.44 & 4.67 & 9.89 & 22.07 & 71.19 \\
&
& BaseShared    & 1.61 & 2.13 & 3.08 & 5.25 & 10.54 & 23.80 & 68.19 \\
&
& PreLRShared   & 1.22 & 1.39 & 1.90 & 2.67 & 5.36 & 12.44 & 27.08 \\
&
& ReBaseShared$_{\mathrm{LP}}$
                & 1.23 & 1.36 & 1.94 & 3.19 & 6.06 & 12.53 & 27.24 \\

\cmidrule(lr){2-10}

&
& \multicolumn{1}{c}{} \\
[-2.2ex]
&
\multirow{8}{*}{TP (tok/s)}
& NonShared     & 176.2 & 263.4 & 402.5 & 594.0 & 674.2 & 767.9 & 775.6 \\
&
& FullShared    & 195.4 & 291.1 & 465.9 & 801.0 & 1249.0 & 1692.5 & 1841.7 \\
&
& DroidSpeak    & 182.1 & 263.2 & 409.2 & 631.8 & 851.6 & 936.9 & 810.0 \\
&
& CacheBlend    & 170.8 & 232.2 & 417.0 & 650.3 & 885.7 & 1042.1 & 933.8 \\
&
& RelayCaching  & 161.7 & 258.8 & 403.2 & 622.0 & 853.5 & 899.7 & 773.9 \\
&
& BaseShared    & 168.8 & 254.6 & 396.0 & 622.1 & 854.4 & 975.5 & 821.0 \\
&
& PreLRShared   & 167.1 & 251.0 & 412.3 & 706.3 & 1144.9 & 1580.9 & 1685.4 \\
&
& ReBaseShared$_{\mathrm{LP}}$
                & 166.2 & 245.2 & 403.9 & 635.1 & 899.5 & 1086.8 & 1068.1 \\

\midrule

\multirow{16}{*}{Ministral-8B}
& \multirow{8}{*}{TTFT (s)}
& NonShared     & 2.03 & 2.63 & 3.83 & 6.86 & 17.52 & 37.65 & 73.55 \\
&
& FullShared    & 1.18 & 1.35 & 1.71 & 2.50 & 4.41 & 9.30 & 20.74 \\
&
& DroidSpeak    & 1.64 & 2.23 & 3.29 & 5.71 & 11.57 & 26.60 & 61.45 \\
&
& CacheBlend    & 1.63 & 2.13 & 2.89 & 3.98 & 10.35 & 21.44 & 60.58 \\
&
& RelayCaching  & 1.79 & 3.19 & 4.29 & 5.12 & 9.86 & 24.83 & 74.96 \\
&
& BaseShared    & 1.65 & 2.19 & 3.20 & 5.46 & 10.76 & 25.63 & 59.34 \\
&
& PreLRShared   & 1.27 & 1.36 & 1.92 & 2.77 & 5.84 & 13.25 & 24.32 \\
&
& ReBaseShared$_{\mathrm{LP}}$
                & 1.25 & 1.33 & 1.88 & 2.74 & 5.95 & 13.03 & 24.68 \\

\cmidrule(lr){2-10}

&
& \multicolumn{1}{c}{} \\
[-2.2ex]
&
\multirow{8}{*}{TP (tok/s)}
& NonShared     & 159.3 & 231.5 & 362.1 & 540.4 & 615.4 & 663.9 & 719.1 \\
&
& FullShared    & 171.0 & 250.9 & 420.2 & 705.4 & 1174.8 & 1540.0 & 1784.5 \\
&
& DroidSpeak    & 161.2 & 249.5 & 360.4 & 574.6 & 780.1 & 861.9 & 833.6 \\
&
& CacheBlend    & 146.0 & 247.4 & 398.3 & 613.3 & 797.6 & 970.7 & 855.7 \\
&
& RelayCaching  & 138.7 & 205.7 & 314.5 & 570.3 & 821.3 & 890.7 & 722.4 \\
&
& BaseShared    & 156.5 & 229.6 & 363.7 & 553.1 & 796.3 & 882.2 & 873.0 \\
&
& PreLRShared   & 149.8 & 227.1 & 371.4 & 634.3 & 1058.7 & 1432.2 & 1639.8 \\
&
& ReBaseShared$_{\mathrm{LP}}$
                & 151.7 & 227.0 & 361.3 & 587.9 & 878.4 & 1043.4 & 1089.5 \\

\bottomrule
\end{tabular}
}
\end{table*}

\newpage
\subsection{Concurrent Server-Side Serving}
\label{app:eff_server}

Section~\ref{sec:exp_eff} presents the concurrent serving trends over request rate, while Table~\ref{tab:eff_server} provides the complete throughput and TTFT percentile results.
The experiments use a fixed $17.3$k-token trajectory on vLLM with a single A100 GPU at request rates ranging from $0.5$ to $16$ QPS.
QPS denotes the agent-call arrival rate, and TTFT percentiles are computed across agent calls, including queueing.
Throughput is the trajectory length divided by the mean sum of agent-call latencies per trajectory, with each DB call completing when both paths finish.
We replay controlled traces with identical agent schedules and token counts across methods, using random token ids; for ReBaseShared, both schedules reconstruct the same span of positions.

Among our methods, PreLRShared maintains the highest per-request throughput and the lowest p50 TTFT across the request-rate sweep, while ReBaseShared$_{\mathrm{DB}}$ also improves both metrics over BaseShared.
The p90 and p99 results show that differences become larger after saturation as queueing accumulates.
In our vLLM implementation, ReBaseShared$_{\mathrm{DB}}$ issues one adapter-free request per turn, covering the same prompt and generated token positions as the adapted path.

\begin{table*}[h]
\centering
\caption{Serving TTFT percentiles in seconds and per-request throughput in tokens per second over request rate with a fixed $17.3$k-token trajectory on a single A100 GPU. ReBaseShared uses DB scheduling, and TP denotes throughput.}
\label{tab:eff_server}
\begin{tabular}{llrrrrrr}
\toprule
\multirow{2}{*}{Metric}
& \multirow{2}{*}{Method}
& \multicolumn{6}{c}{QPS} \\
\cmidrule(lr){3-8}
&
&
0.5 & 1 & 2 & 4 & 8 & 16 \\
\midrule

\multirow{8}{*}{TP (tok/s)}
& NonShared     & 1573 & 1282 & 596  & 70   & 52  & 48 \\
& FullShared    & 2922 & 2910 & 2818 & 1809 & 205 & 164 \\
& DroidSpeak    & 1592 & 1416 & 686  & 78   & 60  & 53 \\
& CacheBlend    & 1939 & 1847 & 1391 & 140  & 104 & 83 \\
& RelayCaching  & 1827 & 1678 & 835  & 93   & 68  & 59 \\
& BaseShared    & 1528 & 1451 & 758  & 114  & 79  & 68 \\
& PreLRShared   & 2506 & 2518 & 2367 & 1438 & 198 & 144 \\
& ReBaseShared$_{\mathrm{DB}}$
                & 2036 & 1972 & 1408 & 182 & 103 & 85 \\

\midrule

\multirow{8}{*}{p50 TTFT (s)}
& NonShared     & 0.37 & 0.38 & 0.43 & 2.85 & 7.23 & 9.66 \\
& FullShared    & 0.05 & 0.04 & 0.05 & 0.06 & 0.50 & 2.48 \\
& DroidSpeak    & 0.34 & 0.33 & 0.40 & 2.74 & 6.87 & 9.21 \\
& CacheBlend    & 0.25 & 0.24 & 0.24 & 1.23 & 5.30 & 7.79 \\
& RelayCaching  & 0.32 & 0.32 & 0.42 & 3.62 & 7.52 & 9.86 \\
& BaseShared    & 0.36 & 0.37 & 0.43 & 3.52 & 7.19 & 9.15 \\
& PreLRShared   & 0.06 & 0.06 & 0.08 & 0.16 & 1.71 & 2.93 \\
& ReBaseShared$_{\mathrm{DB}}$
                & 0.07 & 0.07 & 0.21 & 1.29 & 4.84 & 6.83 \\

\midrule

\multirow{8}{*}{p90 TTFT (s)}
& NonShared     & 0.63 & 0.85 & 1.48 & 5.02 & 13.28 & 17.16 \\
& FullShared    & 0.46 & 0.46 & 0.45 & 0.46 & 0.87 & 3.19 \\
& DroidSpeak    & 0.57 & 0.88 & 1.35 & 5.56 & 12.64 & 16.22 \\
& CacheBlend    & 0.63 & 0.63 & 0.64 & 2.39 & 9.28 & 13.16 \\
& RelayCaching  & 0.77 & 0.77 & 0.99 & 6.65 & 13.88 & 17.91 \\
& BaseShared    & 0.70 & 0.82 & 1.40 & 6.26 & 13.27 & 16.74 \\
& PreLRShared   & 0.50 & 0.49 & 0.50 & 0.51 & 2.27 & 4.74 \\
& ReBaseShared$_{\mathrm{DB}}$
                & 0.50 & 0.54 & 0.89 & 1.92 & 8.25 & 11.80 \\

\midrule

\multirow{8}{*}{p99 TTFT (s)}
& NonShared     & 1.84 & 1.85 & 1.93 & 5.50 & 14.03 & 18.40 \\
& FullShared    & 0.53 & 0.52 & 0.52 & 0.56 & 1.48 & 4.39 \\
& DroidSpeak    & 1.70 & 1.56 & 1.71 & 5.86 & 13.24 & 17.20 \\
& CacheBlend    & 0.92 & 0.93 & 0.93 & 2.97 & 9.74 & 14.29 \\
& RelayCaching  & 0.97 & 0.97 & 1.05 & 7.46 & 15.18 & 18.95 \\
& BaseShared    & 1.49 & 1.49 & 1.68 & 6.78 & 13.57 & 16.97 \\
& PreLRShared   & 0.58 & 0.56 & 0.56 & 0.60 & 2.52 & 4.97 \\
& ReBaseShared$_{\mathrm{DB}}$
                & 0.54 & 0.59 & 1.07 & 2.31 & 9.43 & 13.23 \\

\bottomrule
\end{tabular}
\end{table*}

\newpage
\subsection{LP and DB Comparison}
\label{app:eff_schedule}

Section~\ref{sec:exp_eff} presents LP and DB scheduling for ReBaseShared, while Tables~\ref{tab:eff_schedule_edge} and~\ref{tab:eff_schedule_server} provide the complete numerical results for single-stream and concurrent execution.
Because both schedules construct logically equivalent cache states, this comparison isolates when adapter-free reconstruction is performed.
LP is preferable for single-stream inference because it performs reconstruction as a contiguous post-turn prefill rather than maintaining a second path throughout decoding.
When external tools are invoked, this post-turn prefill can additionally overlap with tool execution.
DB is preferable for concurrent serving because it integrates reconstruction into the continuous serving batch and avoids a separate prefill that delays queued requests.

\begin{table*}[h]
\centering
\caption{LP and DB scheduling for ReBaseShared under single-stream inference on LLaMA-3.1-8B. TP denotes throughput.}
\label{tab:eff_schedule_edge}
\begin{tabular}{llrrrrrrr}
\toprule
Metric & Schedule
& 1.9k & 3.0k & 5.0k & 9.1k & 17.3k & 33.7k & 66.4k \\
\midrule

\multirow{2}{*}{TTFT (s)}
& LP & 1.23 & 1.36 & 1.94 & 3.19 & 6.06 & 12.53 & 27.24 \\
& DB & 1.24 & 1.58 & 2.33 & 3.94 & 7.72 & 17.95 & 49.29 \\

\midrule

\multirow{2}{*}{TP (tok/s)}
& LP & 166 & 245 & 404 & 635 & 900 & 1087 & 1068 \\
& DB & 126 & 189 & 299 & 486 & 723 & 907 & 889 \\

\bottomrule
\end{tabular}
\end{table*}

\begin{table*}[h]
\centering
\caption{LP and DB scheduling for ReBaseShared under concurrent serving with a fixed $17.3$k-token trajectory. TP denotes throughput.}
\label{tab:eff_schedule_server}
\begin{tabular}{llrrrrrr}
\toprule
\multirow{2}{*}{Metric}
& \multirow{2}{*}{Schedule}
& \multicolumn{6}{c}{QPS} \\
\cmidrule(lr){3-8}
&
&
0.5 & 1 & 2 & 4 & 8 & 16 \\
\midrule

\multirow{2}{*}{p50 TTFT (s)}
& LP & 0.06 & 0.06 & 6.97 & 18.26 & 22.67 & 24.79 \\
& DB & 0.07 & 0.07 & 0.21 & 1.29  & 4.84  & 6.83 \\

\midrule

\multirow{2}{*}{TP (tok/s)}
& LP & 1670 & 1371 & 947  & 69  & 47  & 38 \\
& DB & 2036 & 1972 & 1408 & 182 & 103 & 85 \\

\bottomrule
\end{tabular}
\end{table*}

\subsection{Memory Usage}
\label{app:eff_memory}

Section~\ref{sec:exp_eff} summarizes peak GPU memory usage over trajectory length, while Table~\ref{tab:eff_memory} provides the complete numerical results.
The experiment measures LLaMA-3.1-8B-Instruct from $1.9$k to $66.4$k tokens and includes model weights, persistent cache states, and temporary tensors in peak memory.
PreLRShared and ReBaseShared store one shared base cache and compact per-agent LR caches rather than full-dimensional agent-specific KV caches.
During adapter-free reconstruction, temporary hidden states are released after each layer and do not form a second persistent full-dimensional cache.
Consequently, both methods remain close to FullShared and BaseShared, while their memory advantage over selective recomputation increases with trajectory length.

\begin{table}[h]
\centering
\caption{Peak GPU memory usage over trajectory length on LLaMA-3.1-8B.}
\label{tab:eff_memory}
\begin{tabular}{lrrrrrrr}
\toprule
Method
& 1.9k & 3.0k & 5.0k & 9.1k & 17.3k & 33.7k & 66.4k \\
\midrule
NonShared     & 15.69 & 16.07 & 16.83 & 18.33 & 21.35 & 27.39 & 39.47 \\
FullShared    & 15.29 & 15.47 & 15.85 & 16.60 & 18.10 & 21.11 & 27.13 \\
DroidSpeak    & 15.52 & 15.86 & 16.48 & 17.75 & 20.29 & 25.37 & 35.50 \\
CacheBlend    & 15.49 & 15.77 & 16.36 & 17.52 & 19.86 & 24.52 & 33.86 \\
RelayCaching  & 15.46 & 15.73 & 16.28 & 17.38 & 19.59 & 24.01 & 32.84 \\
BaseShared    & 15.30 & 15.49 & 15.88 & 16.66 & 18.22 & 21.34 & 27.58 \\
PreLRShared   & 15.29 & 15.48 & 15.86 & 16.62 & 18.15 & 21.21 & 27.33 \\
ReBaseShared  & 15.29 & 15.49 & 15.87 & 16.63 & 18.16 & 21.23 & 27.34 \\
\bottomrule
\end{tabular}
\end{table}

\newpage
\subsection{Effect of Agent Number}
\label{app:eff_agentn}

Section~\ref{sec:exp_eff} evaluates efficiency with a fixed number of agents while varying trajectory length.
Here, we fix the retrieved context to $8$k tokens per turn and the total execution to eight turns, while increasing the number of agents from $N=4$ to $N=8$ under round-robin execution.
Each additional agent introduces another rank-$r$ down projection and LR cache.
Both settings reach approximately the same $64$k-token trajectory, but increasing $N$ increases the amount of accumulated context that the active agent has not previously processed.
This setup isolates agent-number scalability from total trajectory length.
Table~\ref{tab:eff_agentn} reports the results.

\begin{table}[h]
\centering
\caption{Efficiency and peak GPU memory usage for different numbers of agents on LLaMA-3.1-8B-Instruct using a single A6000 GPU. Each turn adds $8$k tokens over a total of eight turns.}
\label{tab:eff_agentn}
\begin{tabular}{llrr}
\toprule
Metric & Method & $N=4$ & $N=8$ \\
\midrule

\multirow{8}{*}{TTFT (s)}
& NonShared
& 68.36 & 90.05 \\
& FullShared
& 23.98 & 22.73 \\
& DroidSpeak
& 66.57 & 88.89 \\
& CacheBlend
& 48.71 & 72.13 \\
& RelayCaching
& 52.99 & 80.96 \\
& BaseShared
& 58.09 & 87.60 \\
& PreLRShared
& 26.36 & 27.17 \\
& ReBaseShared$_{\mathrm{LP}}$
& 27.86 & 29.76 \\

\midrule

\multirow{8}{*}{Throughput (tok/s)}
& NonShared
& 770 & 569 \\
& FullShared
& 1814 & 1882 \\
& DroidSpeak
& 794 & 582 \\
& CacheBlend
& 1041 & 776 \\
& RelayCaching
& 1003 & 703 \\
& BaseShared
& 835 & 664 \\
& PreLRShared
& 1615 & 1516 \\
& ReBaseShared$_{\mathrm{LP}}$
& 1073 & 1006 \\

\midrule

\multirow{8}{*}{Peak Memory (GB)}
& NonShared
& 31.14 & 39.28 \\
& FullShared
& 25.12 & 25.15 \\
& DroidSpeak
& 28.83 & 34.73 \\
& CacheBlend
& 27.99 & 32.98 \\
& RelayCaching
& 27.45 & 31.86 \\
& BaseShared
& 25.56 & 26.09 \\
& PreLRShared
& 25.38 & 25.87 \\
& ReBaseShared$_{\mathrm{LP}}$
& 25.36 & 25.62 \\
\bottomrule
\end{tabular}
\end{table}

From $N=4$ to $N=8$, the TTFT of PreLRShared and ReBaseShared$_{\mathrm{LP}}$ increases by only $3\%$ and $7\%$, respectively, compared with increases of $32$--$53\%$ for NonShared, BaseShared, and selective recomputation.
Their throughput also decreases by approximately $6\%$, while these baselines decrease by $20$--$30\%$.
With more agents, these baselines process or recompute more accumulated context that the active agent has not previously processed.
PreLRShared avoids this processing by constructing the LR caches for all agents when each segment is first processed.
ReBaseShared$_{\mathrm{LP}}$ retains this precomputation and reconstructs each newly added segment once with the adapter-free backbone.
Because the total amount of newly added context remains fixed, its reconstruction cost does not scale directly with $N$.
LP reconstruction remains part of wall-clock completion time, explaining the lower throughput of ReBaseShared$_{\mathrm{LP}}$ relative to PreLRShared despite their similar TTFT scaling.

Memory usage follows a similar scaling trend.
PreLRShared and ReBaseShared store one shared base cache and compact per-agent LR caches, limiting their peak memory increase to approximately $1$--$2\%$ as $N$ doubles.
In contrast, selective recomputation retains full-dimensional agent-specific states for recomputed tokens or layers, increasing peak memory by $16$--$20\%$, while NonShared increases by $26\%$ because each agent maintains a full KV cache.
Thus, the computation and memory costs of PreLRShared and ReBaseShared depend primarily on the total newly added context rather than the interval between agent turns.

\newpage
\section{LoRA Analysis}
\label{app:lora}

\subsection{LoRA Rank}
\label{app:lora_rank}

Section~\ref{sec:exp_setup} uses $r=8$ as the default LoRA rank.
This appendix examines its effect on benchmark accuracy and serving efficiency.
Table~\ref{tab:lora_rank_acc} reports HotpotQA accuracy on Ministral-8B-Instruct and includes RelayCaching as a representative selective recomputation baseline.
Table~\ref{tab:lora_rank_tp} reports single-stream throughput at a fixed $33.7$k-token trajectory on LLaMA-3.1-8B-Instruct.

\renewcommand{\arraystretch}{0.9}
\begin{table}[h]
\centering
\caption{HotpotQA accuracy (\%) across LoRA ranks on Ministral-8B-Instruct.}
\label{tab:lora_rank_acc}
\begin{tabular}{lrrrr}
\toprule
Method & $r=4$ & $r=8$ & $r=16$ & $r=32$ \\
\midrule
NonShared
& 34.72 & 36.72 & 36.70 & 36.80 \\
RelayCaching
& 32.95 & 34.63 & 34.47 & 34.20 \\
PreLRShared
& 33.51 & 34.77 & 34.67 & 34.53 \\
ReBaseShared
& 34.41 & 36.08 & 35.98 & 35.81 \\
\bottomrule
\end{tabular}
\end{table}
\renewcommand{\arraystretch}{1.0}

Accuracy increases from $r=4$ to $r=8$ for all methods and changes little at higher ranks.
Increasing the rank beyond $8$ also does not reduce the accuracy gap introduced by KV cache sharing, indicating that this gap is not caused by insufficient adapter capacity.
ReBaseShared remains within $1.0\%$ points of NonShared across $r=8$--$32$, with its accuracy varying by less than $0.3\%$ points.
These results support $r=8$ as the default rank without sacrificing accuracy.

\begin{table}[h]
\centering
\caption{Single-stream throughput in tokens per second across LoRA ranks at a fixed $33.7$k-token trajectory on LLaMA-3.1-8B-Instruct.}
\label{tab:lora_rank_tp}
\begin{tabular}{lrrrr}
\toprule
Method & $r=8$ & $r=16$ & $r=32$ & $r=64$ \\
\midrule
NonShared
& 768 & 762 & 756 & 750 \\
FullShared
& 1693 & 1696 & 1694 & 1690 \\
DroidSpeak
& 937 & 927 & 921 & 913 \\
CacheBlend
& 1042 & 1049 & 1042 & 1033 \\
RelayCaching
& 900 & 902 & 891 & 878 \\
BaseShared
& 976 & 967 & 958 & 953 \\
PreLRShared
& 1581 & 1578 & 1569 & 1547 \\
ReBaseShared$_{\mathrm{LP}}$
& 1087 & 1083 & 1071 & 1048 \\
\bottomrule
\end{tabular}
\end{table}

Throughput changes modestly as the rank increases from $8$ to $64$.
PreLRShared and ReBaseShared$_{\mathrm{LP}}$ decrease by $2\%$ and $4\%$, respectively, because the cost of LR cache construction and Flash-LoRA-Attention increases with $r$.
Nevertheless, PreLRShared remains within $9\%$ of FullShared across all ranks.
Together with the accuracy saturation above, these results show that $r=8$ provides a suitable balance between adapter capacity and LR cache processing overhead.

\newpage
\subsection{LoRA Projection Configuration}
\label{app:lora_qkvo}

Section~\ref{sec:exp_setup} applies LoRA to the query and value projections using QV adaptation.
This appendix additionally considers QKVO adaptation with $r=4$, which has the same number of LoRA parameters as QV adaptation with $r=8$.
This parameter-matched setting isolates the effect of adapting additional attention projections from that of increasing the adapter size.
Table~\ref{tab:lora_qkvo_eff} reports single-stream TTFT and throughput over trajectory length on LLaMA-3.1-8B-Instruct.

\begin{table*}[h]
\centering
\caption{Single-stream TTFT in seconds and throughput in tokens per second under QKVO adaptation with $r=4$ on LLaMA-3.1-8B-Instruct. This configuration matches the LoRA parameter count of QV adaptation with $r=8$, and TP denotes throughput.}
\label{tab:lora_qkvo_eff}
\resizebox{\linewidth}{!}{
\begin{tabular}{llrrrrrrr}
\toprule
Metric & Method
& 1.9k & 3.0k & 5.0k & 9.1k & 17.3k & 33.7k & 66.4k \\
\midrule

\multirow{8}{*}{TTFT (s)}
& NonShared
& 1.81 & 4.16 & 4.39 & 6.45 & 12.91 & 27.01 & 68.73 \\
& FullShared
& 1.45 & 1.45 & 2.00 & 2.71 & 5.23 & 10.18 & 24.62 \\
& DroidSpeak
& 1.76 & 2.28 & 3.38 & 5.82 & 11.49 & 25.46 & 65.37 \\ 
& CacheBlend
& 2.11 & 2.15 & 3.86 & 4.12 & 7.48 & 19.73 & 59.78 \\
& RelayCaching
& 3.16 & 3.48 & 3.44 & 6.37 & 9.14 & 24.58 & 71.56 \\
& BaseShared
& 1.80 & 2.36 & 3.26 & 5.63 & 10.99 & 26.05 & 70.94 \\ 
& PreLRShared
& 1.90 & 2.50 & 3.07 & 4.16 & 7.74 & 16.50 & 32.55 \\
& ReBaseShared$_{\mathrm{LP}}$
& 2.15 & 2.93 & 3.35 & 6.20 & 10.15 & 22.53 & 55.94 \\

\midrule

\multirow{8}{*}{TP (tok/s)}
& NonShared
& 136.3 & 170.7 & 282.8 & 464.3 & 661.3 & 784.1 & 729.7 \\
& FullShared
& 139.9 & 201.9 & 327.0 & 573.8 & 936.1 & 1288.7 & 1416.1 \\
& DroidSpeak
& 136.7 & 191.1 & 300.1 & 479.2 & 697.9 & 814.0 & 753.9 \\
& CacheBlend
& 133.4 & 192.8 & 291.5 & 526.9 & 834.5 & 944.0 & 809.3 \\
& RelayCaching
& 124.3 & 177.6 & 298.9 & 466.2 & 772.7 & 831.1 & 707.7 \\
& BaseShared
& 136.6 & 190.5 & 301.6 & 485.6 & 713.7 & 799.1 & 715.2 \\
& PreLRShared
& 135.4 & 188.6 & 305.6 & 525.7 & 824.2 & 1037.9 & 1211.2 \\
& ReBaseShared$_{\mathrm{LP}}$
& 133.1 & 183.6 & 300.5 & 470.3 & 739.3 & 875.3 & 849.0 \\
\bottomrule
\end{tabular}
}
\end{table*}

Unlike QV adaptation, QKVO adaptation makes the key cache agent-specific and introduces an additional key LR cache.
The key-side adapter contribution must be expanded from rank $r$ before applying RoPE, so the associativity-based reordering used by Flash-LoRA-Attention for the value LR cache does not directly apply.
QKVO therefore incurs additional LR cache computation relative to the default QV setting.
Unlike architecture-constrained methods that restrict adapter placement to preserve identical KV caches, PReCache supports LoRA adaptation in the key and value projections~\citep{woo2026icarus,woo2026prefillshare}.

Nevertheless, the long-context efficiency trend remains consistent with the main experiments.
At the longest trajectory, PreLRShared achieves the highest efficiency among the methods other than FullShared and remains closest to FullShared, while ReBaseShared$_{\mathrm{LP}}$ retains an advantage over the remaining baselines despite its adapter-free reconstruction.
BaseShared is affected more strongly because it processes the accumulated context with the current agent's backbone and additionally constructs the key LR cache.
Consequently, at $66.4$k tokens, BaseShared no longer improves TTFT or throughput over NonShared, whereas PreLRShared and ReBaseShared$_{\mathrm{LP}}$ retain both improvements.
Together with the lower accuracy of parameter-matched QKVO adaptation reported by LRAgent~\citep{jeon2026lragent}, these results support QV as the default configuration while showing that PReCache remains effective when the key projection is also adapted.

\end{document}